\documentclass[letterpaper, 10 pt, conference]{ieeeconf}  

\IEEEoverridecommandlockouts                              

\usepackage{graphicx} 
\usepackage{booktabs,longtable,makecell, array}
\newcolumntype{L}[1]{>{\raggedright\arraybackslash}p{#1}}
\usepackage{multirow}
\usepackage{afterpage}
\usepackage{subfig}
\usepackage{amsmath}
\usepackage{amsfonts}
\usepackage{array}
\usepackage{tabularx}
\usepackage{booktabs}
\usepackage{subcaption}
\usepackage{url}
\usepackage[font=footnotesize, labelfont=footnotesize, textfont=footnotesize]{caption}
\usepackage{hyperref}
\usepackage{adjustbox}

\makeatletter
\newcommand{\setcaptype}[1]{\def\@captype{#1}}
\makeatother

\newsavebox{\tempbox}

\title{\LARGE \bf
Scalable Multi-Objective Robot Reinforcement Learning through Gradient Conflict Resolution
}
\author{Humphrey Munn$^{1,2}$, Brendan Tidd$^{2}$, Peter B{\"o}hm$^{1,2}$, Marcus Gallagher$^{1}$, David Howard$^{2}$
\thanks{$^{1}$Humphrey Munn, Peter B{\"o}hm, and Marcus Gallagher are with the School of Electrical Engineering and Computer Science,
        University of Queensland, QLD. 4072, Australia
        {\{\tt\small h.munn,p.bohm,marcusg\}}{\tt\small@uq.edu.au}}
\thanks{$^{2}$Humphrey Munn, Brendan Tidd, Peter B{\"o}hm, and David Howard are with 
{\tt\small \{humphrey.munn,brendan.tidd,peter.bohm,\newline david.howard\}}
        {\tt\small@data61.csiro.au}}}%

\begin{document}

\maketitle
\savebox{\tempbox}{\begin{minipage}{\textwidth}
\centering
    \vspace{-0.6cm}
    \includegraphics[width=1\linewidth]{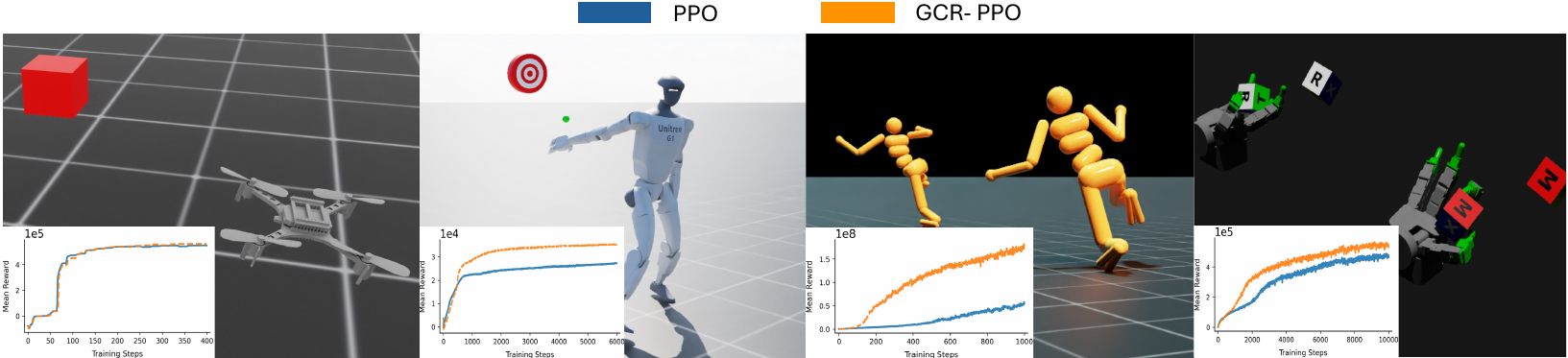}
    \captionof{figure}{\footnotesize Comparison of GCR-PPO and massively parallel PPO on four select tasks tested: Quadcopter, Full-Body Throwing (multi objective), Humanoid Running (multi objective: gait length, energy usage, arm span, base height), and Allegro-Cube. Both methods perform similarly on the Quadcopter, where gradient conflict is low. GCR-PPO achieves substantial gains on the other three tasks, which show high gradient conflict. Overall, GCR-PPO mitigates conflicts between objectives and scales more favourably than standard parallelised PPO.}
   \label{fig:heroshot}
    \vspace{-0.6cm}
\end{minipage}}
\begin{figure}[t]
\rlap{\usebox\tempbox}
\end{figure}
\afterpage{\begin{figure}[t]
\rule{0pt}{\dimexpr \ht\tempbox+\dp\tempbox}

\end{figure}}

%

\begin{abstract}

Reinforcement Learning (RL) robot controllers usually aggregate many task objectives into one scalar reward. While large-scale proximal policy optimisation (PPO) has enabled impressive results such as robust robot locomotion in the real world, many tasks still require careful reward tuning and are brittle to local optima. Tuning cost and sub-optimality grow with the number of objectives, limiting scalability. Modelling reward vectors and their trade-offs can address these issues; however, multi-objective methods remain underused in RL for robotics because of computational cost and optimisation difficulty. In this work, we investigate the conflict between gradient contributions for each objective that emerge from scalarising the task objectives. In particular, we explicitly address the conflict between task-based rewards and terms that regularise the policy towards realistic behaviour. We propose GCR-PPO, a modification to actor-critic optimisation that decomposes the actor update into objective-wise gradients using a multi-headed critic and resolves conflicts based on the objective priority. Our methodology, GCR-PPO, is evaluated on the well-known IsaacLab manipulation and locomotion benchmarks and additional multi-objective modifications on two related tasks. We show superior scalability compared to parallel PPO (p = 0.04), without significant computational overhead. We also show higher performance with more conflicting tasks. GCR-PPO improves on large-scale PPO with an average improvement of 9.5\% (Symmetric Percentage Change), with high-conflict tasks observing a greater improvement \footnote{The code is available at https://github.com/humphreymunn/GCR-PPO.}.

\end{abstract}


\section{INTRODUCTION}

As robotics continues to find application in increasingly complex real-world tasks, scalable Reinforcement Learning (RL) pipelines become increasingly important. Scalability issues persist across various parts of robotic RL \cite{tang2025deep,farooq2024survey}, for example in \textit{sample efficiency} (as the robot must repeatedly interact with its environment to iteratively tune its behaviour), and in \textit{behavioural complexity}, which refers to learning larger numbers of tasks or objectives to achieve more capable and flexible behaviours.  Issues remain with both, but a combination of (i) recent development of reasonably accurate and fast parallel simulators \cite{rudin2022learning, collins2021review} to address the former problem, and (ii) a relative lack of literature addressing the latter issue, behavioural complexity is arguably the more relevant.

Behavioural complexity takes two primary forms - multi-task and multi-objective.  Multi-task can be thought of as, for example, a `kitchen robot' that can boil a kettle, open the fridge, cook, and do the washing up \cite{yu2020meta}. Each task must be explicitly defined, atomic and separated in time.  Multi-objective addresses the challenge of simultaneously balancing multiple requirements, which may change through time, and thus provides an opportunity for emergent behaviour without explicitly defining tasks.  Adapting to balance larger numbers of objectives allows for more complex behaviours to be realised and is therefore a precursor to successful deployment of useful general-purpose robotics solutions in, e.g., the kitchen or a laboratory.  However, for the state of the art in multi-objective RL for robotics, issues including the inability to effectively handle conflicts between objectives and the poor sample efficiency/generalisation across objectives mean scalability and complexity are limited.

We directly address this issue by focusing on \textit{scalable multi-objective learning}, which practically involves combining multiple reward signals into a single policy without disrupting the overall optimisation process, allowing us to encourage the development of more complex behaviours by representing larger numbers of objectives in the reward function.  We develop a solution based on gradient repair \cite{yu2020gradient} for actor-critic methods, which we call Gradient Conflict Resolution or GCR. In this work, we integrate gradient conflict resolution (GCR) into Proximal Policy Optimisation (PPO), referring to the resulting algorithm as GCR-PPO.

GCR-PPO provides a scalable approach for reinforcement learning agents to balance and learn from multiple reward signals simultaneously through conflict resolution, thereby preventing important objectives from being lost in scalarisation and leading to more nuanced and capable behaviours. Contrasting previous approaches~\cite{roijers2013survey}, GCR-PPO explicitly handles conflicting objectives, which are ubiquitous in real-world scenarios and remain a central challenge for modern RL, while scaling effectively with increasing numbers of objectives.

Our key novel contributions are:

\begin{itemize}
    \item Prioritisation-based gradient resolution, extending PCGrad\cite{yu2020gradient} to respect the robotics-specific distinction between task objectives and regularisers.
    \item A multi-head critic for efficient per-reward advantage estimation.
    \item Large-scale comparisons against massively parallel GPU-based PPO on challenging robotic RL benchmarks, including custom multi-objective tasks.
\end{itemize}

Through a series of experiments, we show that GCR-PPO scales better than PPO with the number of objectives, enabling more complex behaviours. For example, Humanoid Running discovers a stylised fast gait—long strides at a low base height—. In contrast, massively parallel PPO typically solves only the fundamental running objective and fails to realise the stylised variant. Across 13 IsaacLab tasks, GCR-PPO achieves an average +4.6\% symmetric percentage change (SPC) over PPO, with gains increasing in tasks exhibiting higher gradient conflict (Spearman $\rho$ = 0.736, p = 0.0041). On two custom multi-objective suites, it reaches up to +542\% SPC relative to massively parallel PPO for specific Humanoid configurations. A consistent trend emerges: as more objectives are added, the PPO baseline is prone to poor convergence points that optimise a single term rather than the joint objective set, whereas GCR-PPO more reliably identifies and optimises a feasible subset of objectives jointly, reducing conflicts over training.

\section{RELATED WORK}

Managing and integrating potentially conflicting behavioural goals is a central challenge in robot control, where the system must still achieve its overall objective \cite{althaus2003behavior,pirjanian2000multiple}. A longstanding issue in robotic reinforcement learning is the construction of objective functions that enable complex, capable, and safe behaviours \cite{gu2025safe,huang2022constrained}. The gap lies between efficient single-objective RL, which optimises tasks using a single scalar reward, and the richer reward geometries required to simultaneously optimise for task success while regularising behaviour to be realistic and safe. Addressing this gap is necessary for scalable RL with many objectives, particularly in more complex tasks.


\subsection{Classical Control and Task Prioritisation}
Early solutions to conflicting objectives in robotics were developed in classical control, most notably prioritised control and stack-of-tasks methods \cite{siciliano2009robotics, khatib1987whole}. These approaches project lower-priority objectives into the null space of higher-priority ones, ensuring that secondary goals never compromise critical tasks (e.g., balance or obstacle avoidance). They have been successfully applied in humanoid and whole-body control, where accurate system models and strict task hierarchies are available. However, these methods depend on exact system models and manually specified task hierarchies, which limit their applicability in modern reinforcement learning. 

In RL, objectives are encoded implicitly through reward terms rather than explicit priorities, and interaction is often model-free with noisy gradient feedback. RL is therefore preferred for scaling to high-dimensional, uncertain environments where rigid model-based prioritisation is impractical \cite{kober2013reinforcement}. Nevertheless, resolving conflicts through projection has influenced more recent gradient-based methods \cite{yu2020gradient,guangyuanrecon,yang2024representation} in learning-based control, and underpins our approach to prioritising objectives defined by reward terms.

\subsection{Single-Objective Reward Combination in Reinforcement Learning}
In most reinforcement learning for robotics, multiple reward terms are combined into a single scalar objective, and standard algorithms such as PPO \cite{schulman2017proximal} with generalised advantage estimation (GAE) \cite{schulman2016gae} are applied to stabilise training. While effective for many tasks, this approach treats all objectives as fully aligned, ignoring potential conflicts. 
Methods like PopArt \cite{hessel2019multi} adaptively rescale value predictions to address scale imbalance, but they do not mitigate interference between competing reward signals. 
Similarly, auxiliary-task methods such as UNREAL \cite{jaderberg2017unreal} and distributed frameworks like IMPALA \cite{espeholt2018impala} enhance sample efficiency and representation learning. Yet, they remain complementary rather than providing a mechanism for resolving conflicting objectives. 
Prior work on reward shaping \cite{ng1999policy} further demonstrates the importance of reward design, but none explicitly address the gradient-level conflicts that arise when multiple objectives must be satisfied simultaneously.

\subsection{Multi-Objective Reinforcement Learning}
Multi-objective reinforcement learning (MORL) \cite{huang2022constrained} extends the standard RL setting to optimise multiple, potentially conflicting objectives simultaneously \cite{roijers2013survey,vamplew2011empirical}. A typical formulation seeks policies that approximate the Pareto front, i.e., the set of policies that are not dominated in all objectives. 
Exact Pareto-optimal methods \cite{vamplew2011empirical} are attractive because they provide principled guarantees but are computationally intractable in high-dimensional problems. 
Approximations—such as computing convex combinations of objectives or using variants of the Multiple Gradient Descent Algorithm (MGDA) \cite{desideri2012multiple} remain costly in high-dimensional settings such as robotic RL or with many objectives, as they require repeated optimisation of vector-valued value functions or constrained sub-problems.

Many MORL methods adopt scalarisation strategies to improve scalability, where a weighted sum of objectives is optimised \cite{van2014multi}. While computationally efficient, scalarisation hides conflicts and requires carefully chosen weights, which may not generalise across tasks or environments. Other approaches explore decomposition methods, such as learning multiple value functions for each objective \cite{mossalam2016multi}, but these remain limited by sample complexity and optimisation overhead.
\subsection{Multi-Task Reinforcement Learning}
Multi-task reinforcement learning (MTRL) aims to train a single policy to solve multiple tasks, typically by sharing representations across tasks \cite{vithayathil2020survey}. 
While multi-objective RL (MORL) focuses on optimising several reward terms within one task, MTRL treats each task as a separate objective instead. Both settings face the same optimisation challenge: gradients from different objectives may conflict, leading to destructive interference during training.   

Recent methods take heuristic approaches: PCGrad \cite{yu2020gradient} projects gradients to reduce interference and shows substantial empirical gains, though without theoretical convergence guarantees. Conflict-averse gradient descent (CAGrad) \cite{liu2021conflict} adds convergence guarantees at the cost of greater computational complexity. In contrast, our approach extends the lightweight PCGrad framework with priority-based projections tailored to the robotics setting, where task objectives and regularisers must be distinguished.

\subsection{Connections Between Multi-Task and Multi-Objective RL}
In multi-task reinforcement learning, gradients from several distinct tasks are aggregated to update a single policy \cite{vithayathil2020survey}. 
Additive-reward RL can be seen as a related formulation: summing reward components into a scalar corresponds to optimising multiple objectives, but with all components sharing the same state–action trajectories rather than distinct task distributions. 
This difference aside, both settings face the same optimisation challenges, notably destructive interference from conflicting gradients. In additive-reward RL, these conflicts are often hidden by scalarisation.

Recent multi-objective RL literature highlights how scalarisation masks such challenges. Vamplew et al. \cite{vamplew2024value} show that scalarisation can lead to suboptimal policies due to value function interference when heterogeneous rewards are collapsed into a single utility. Dann et al. \cite{dann2023reinforcement} demonstrate the benefits of treating multiple rewards explicitly rather than averaging them into a single signal.

A related work is CoMOGA \cite{kim2024conflict}, which frames CMORL updates as constrained sub-problems to aggregate gradients conflict-free with safety guarantees, yielding tabular guarantees and Pareto coverage. In contrast, we address single-task additive rewards in PPO, resolving conflicts with lightweight priority-based PCGrad projections aimed at scalability and on-policy practicality rather than constrained Pareto sets.

Building on these insights, we explicitly decompose the scalar objective into per-component losses and gradients using a multi-head critic architecture. This decomposition makes hidden conflicts observable and allows gradient-surgery techniques from multi-task optimisation (e.g., projection-based conflict resolution) to be applied in the multi-objective setting.


\section{METHOD}
\begin{figure}
    \centering
    \includegraphics[width=0.99\linewidth]{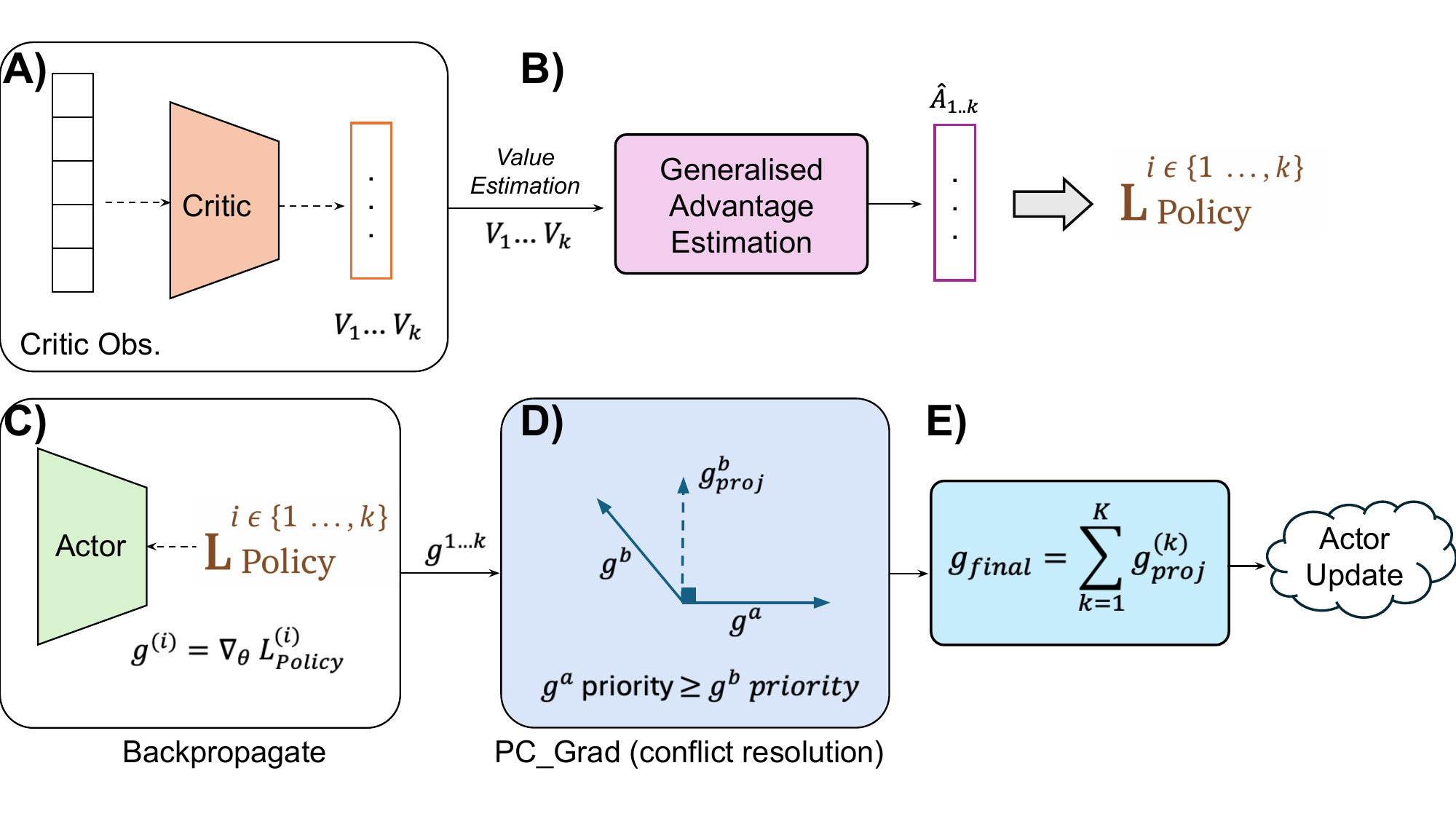}
    \captionof{figure}{\footnotesize 
GCR-PPO assumes an additive reward decomposition $r = r_0 + \dots + r_k$, each term obtains a surrogate loss, and resolves gradient conflicts for scalability to multi-objective settings. 
(A) A multi-headed critic estimates values for each reward term. 
(B) These are processed with GAE and standardised (preserving relative scale) to yield per-term advantages. 
(C) Surrogate losses are computed, and (D) their gradients are projected with PCGrad so conflicts do not weaken higher-priority terms. 
(E) The projected gradients are aggregated into a single update for the actor.}
    \label{fig:fig1}
    \vspace{-0.5cm}
\end{figure}

We adapt the publicly available RSL-RL implementation of proximal policy optimisation (PPO) \cite{rudin2022learning} for our GCR-PPO implementation. All hyperparameters are identical to the default RSL-RL IsaacLab configurations per task, except for the entropy coefficient, which is tuned as described in Section \ref{eval:isaac}.  

\subsection{Reinforcement Learning Formulation}

We consider a discounted Markov Decision Process (MDP)
\[
\mathcal{M} = (\mathcal{S}, \mathcal{A}, P, r, \gamma),
\]
with state space $\mathcal{S}$, action space $\mathcal{A}$, transition kernel $P(s_{t+1}\!\mid\! s_t,a_t)$, reward function $r:\mathcal{S}\times\mathcal{A}\to\mathbb{R}$, and discount factor $\gamma\in(0,1)$. A stochastic policy $\pi_\theta(a_t\!\mid\!s_t)$ maximises the expected discounted return
\[
J(\pi_\theta)=\mathbb{E}_{\pi_\theta}\!\Big[\sum_{t=0}^{\infty}\gamma^{t} r_t\Big].
\]

\noindent We assume an additive per-step reward

\[
r(s_t,a_t)=\sum_{k=1}^{K} r(s_t,a_t)^{(k)},
\]
where $r_t^{(k)}$ are reward components specified by the practitioner. This construction is common in robotics, often including terms for task completion or success, and regularising terms to control efficiency. 

We partition the reward components into two disjoint categories via index sets $\mathcal{I}_{\mathcal{T}}$ (task-based) and $\mathcal{I}_{\mathcal{R}}$ (regulariser-based), with
\[
\mathcal{I}_{\mathcal{T}}\cup\mathcal{I}_{\mathcal{R}}=\{1,\dots,K\}.\qquad
\]
Task-based components reflect objective attainment (e.g., forward velocity tracking, goal proximity, task success), while regulariser components encourage stability or efficiency (e.g., torque penalties, joint-limit costs). This separation enables our method to limit task deterioration due to regularising terms. 

\noindent For each component, define the discounted return
\[
G_t^{(k)}=\sum_{t'=t}^{\infty}\gamma^{\,t'-t} r_{t'}^{(k)}  .
\]
\noindent Then the total return and objective decompose as:
\[
G_t=\sum_{k=1}^{K}G_t^{(k)},\qquad
J(\pi_\theta)=\sum_{k=1}^{K} J^{(k)}(\pi_\theta),\]

\noindent \quad\text{where}\quad \[J^{(k)}(\pi_\theta)=\mathbb{E}_{\pi_\theta}\!\big[G_0^{(k)}\big].
\]

This formulation yields a value function for each reward component. With a minor adaptation of generalised advantage estimation (GAE) \cite{schulman2015high} (Section \ref{method:gae}), we compute component-wise advantages and the corresponding surrogate-policy gradients per objective. These gradients enable the gradient-resolution mechanisms introduced in Section \ref{method:grs}.

\subsection{Multi-head critic and component-wise advantages}\label{method:gae}

We parameterise a vector-valued critic $\mathbf{V}_\phi:\mathcal{S}\to\mathbb{R}^K$ with $k$-th head $V_\phi^{(k)}(s)$ predicting the discounted value of the $k$-th component $r^{(k)}$, i.e.
\[
V_\phi^{(k)}(s_t)\approx \mathbb{E}\!\left[\sum_{\ell=0}^{\infty}\gamma^{\ell}\, r_{t+\ell}^{(k)}\;\middle|\;s_t\right],\qquad k=1,\dots,K.
\]
We train all heads jointly by least squares against bootstrapped per-component targets $\hat G_t^{(k)}$:
\[
\mathcal{L}_V(\phi)=\tfrac{1}{2}\,\mathbb{E}\!\left[\sum_{k=1}^{K}\big(\hat G_t^{(k)}-V_\phi^{(k)}(s_t)\big)^2\right].
\]

Component-wise Generalised Advantage Estimation (GAE) is computed with the per-component temporal-difference residuals
\[
\delta_t^{(k)}=\tilde r_t^{(k)}+\gamma(1-d_t)\,V_\phi^{(k)}(s_{t+1})-V_\phi^{(k)}(s_t),
\]
where $d_t\in\{0,1\}$ indicates termination. For a trajectory segment of length $T$, the component-wise advantages are
\[
A_t^{(k)}=\sum_{\ell=0}^{T-1-t}\big(\gamma\lambda\big)^{\ell}\!\left(\prod_{j=0}^{\ell-1}(1-d_{t+j})\right)\delta_{t+\ell}^{(k)},
\qquad k=1,\dots,K,
\]
and the corresponding TD($\lambda$) targets used in $\mathcal{L}_V$ are
\[
\hat G_t^{(k)}=A_t^{(k)}+V_\phi^{(k)}(s_t).
\]
Stacking across components yields the advantage vector $\mathbf{A}_t=[A_t^{(1)},\dots,A_t^{(K)}]^\top$. 

\subsection{Advantage normalisation strategy}

Directly summing the raw per-component advantages $\{A_t^{(k)}\}_{k=1}^K$ can lead to imbalance, since different reward terms often have very different magnitudes. 
Standard PPO implementations—such as Stable Baselines3 \cite{raffin2021stable}, RSL\_RL \cite{RSLRL2025}, and skrl \cite{serrano2023skrl}—address this by standardising advantages within each batch to unit variance. However, applying this procedure independently to each component would eliminate their relative scale in the multi-component setting. To avoid this, we adopt a normalisation strategy that preserves inter-component ratios while ensuring the combined advantage has unit variance.

We centre each component advantage and apply a global scale so the summed advantage has unit variance while preserving component ratios:
\begin{equation}
\setlength{\arraycolsep}{2pt}
\begin{alignedat}{2}
\hat{\mathbf A}_t &= \tfrac{\mathbf A_t-\mu}{\sqrt{\mathbf 1^\top C \mathbf 1+\varepsilon}},\ &
\mu &= \tfrac1N\sum_{t=1}^N \mathbf A_t,\\
C &= \tfrac1{N-1}\sum_{t=1}^N (\mathbf A_t-\mu)(\mathbf A_t-\mu)^\top,
\end{alignedat}
\end{equation}

where $N$ is the total number of samples in the batch. By construction, $\mathrm{Var}\!\left(\sum_{k=1}^K \tilde A_t^{(k)}\right)=1$. This property provides a stable baseline for subsequent policy-gradient updates.

\subsection{Gradient resolution strategy}\label{method:grs}
We use the standard PPO clipped surrogate for each component with the same clipping coefficient $\epsilon$ as the baseline. With importance ratio $\rho_t=\pi_\theta(a_t\mid s_t)/\pi_{\theta_{\text{old}}}(a_t\mid s_t)$ and normalised advantage $\hat A_t^{(k)}$, the per-component loss is
\begin{equation}
\mathcal{L}^{(k)}_{\text{policy}}(\theta)
= -\,\mathbb{E}\!\left[
\min\!\big(\rho_t\,\hat A_t^{(k)},\;
\operatorname{clip}(\rho_t,\,1-\epsilon,\,1+\epsilon)\,\hat A_t^{(k)}\big)
\right].
\end{equation}
We found that per-component clipping with the same threshold as the baseline (typically 0.2) works well in practice. The default learning-rate scheduler adapts to maintain a target KL divergence, further curbing destructive policy updates. 

The gradient of each loss,
\begin{equation}
\mathbf{g}^{(k)} \;=\; \nabla_{\boldsymbol{\theta}}\, \mathcal{L}^{(k)}(\boldsymbol{\theta}),
\end{equation}
represents the update direction associated with the $k$-th reward component. With multiple objectives, these gradient vectors can conflict—i.e., $\mathbf{g}^{(i)\top}\mathbf{g}^{(j)} < 0$ (negative cosine similarity)—so improving one objective can decrease another.

We use the Projected Conflicting Gradient (PCGrad) algorithm \cite{yu2020gradient} to resolve conflicts that are detrimental to task success. Given two component gradients $\mathbf{g}^{(i)}$ and $\mathbf{g}^{(j)}$, if their inner product is negative ($\mathbf{g}^{(i)\top}\mathbf{g}^{(j)}<0$), we project the \emph{lower-priority} gradient onto the orthogonal complement of the higher-priority one:
\[
\mathbf{g}^{(i)} \;\leftarrow\; \mathbf{g}^{(i)} \;-\; 
\frac{\mathbf{g}^{(i)\top}\mathbf{g}^{(j)}}{\|\mathbf{g}^{(j)}\|^{2}}\,\mathbf{g}^{(j)}.
\]
This removes only the conflicting component, preserving non-conflicting directions and ensuring higher-priority objectives are never weakened. Unlike the original PCGrad, which treats all objectives symmetrically, our variant enforces priority by making the projection asymmetric: lower-priority gradients are adjusted, while higher-priority ones remain unchanged.

Priority is determined by reward type: 
\begin{itemize}
    \item \textbf{Task-based gradients} take precedence. If a conflict occurs between a task-based and a regulariser component, the regulariser gradient is projected onto the task-based direction.  
    \item \textbf{Task--task conflicts} are handled symmetrically with PCGrad, ensuring neither dominates completely.  
    \item \textbf{Regulariser--regulariser conflicts} are also handled symmetrically with PCGrad, as they do not imply conflict with the task. 
\end{itemize}

The final policy update direction is the sum of the adjusted gradients
\[
\mathbf{g_{\text{final}}} = \sum_{k=1}^K \mathbf{g^{(k)}_{\text{proj}}}.
\]
This scheme ensures that task-related objectives drive learning, while regularisers act as soft constraints. Non-deterministic projection maintains diversity of gradient directions across mini-batches, encouraging stable yet task-focused policy improvement.

\section{EXPERIMENTS AND RESULTS}

To assess the effectiveness of GCR-PPO in highly multi-objective settings, we evaluate it on three sets of problems. The first is the \textbf{IsaacLab Task Suite} \cite{IsaacLabDocs2025}, which provides a diverse collection of challenging and realistically deployable tasks. We selected this suite for its breadth and the presence of several environments with many ($\geq$10) reward terms. The second group of evaluations considers extensions of two multi-objective whole-body control problems on a high DoF humanoid robot, \textbf{Full-Body Throwing} \cite{munn2024whole} and \textbf{Humanoid Running} \cite{IsaacLabDocs2025}. We introduce multiple style-based objectives as additional reward components for each, using varying combinations to generate a wide range of multi-objective tasks. This design allows us to test the robustness of GCR-PPO across benchmark and custom problem variations for 213 unique tasks.
\newcommand{\ms}[2]{\ensuremath{#1\,\pm\,#2}}

\begin{table*}[t!]
\vspace{0.2cm}
\centering
\caption{Per-task comparison of GCR-PPO against the PPO baseline. SPC denotes Symmetric Percent Change, and Avg. conflict is the mean number of conflicting component pairs per run (angle $>$ 90°) measured during GCR-PPO training, averaged over seeds.}\label{tab:task-comparison}
{\footnotesize                 
\setlength{\tabcolsep}{7pt}    
\renewcommand{\arraystretch}{1.15} 

\begin{tabular}{@{}p{0.20\textwidth} >{\raggedleft\arraybackslash}p{0.18\textwidth} >{\raggedleft\arraybackslash}p{0.18\textwidth} >{\centering\arraybackslash}p{0.15\textwidth} >{\raggedleft\arraybackslash}p{0.08\textwidth} >{\raggedleft\arraybackslash}p{0.08\textwidth}@{}}

\toprule
\makecell[l]{\textbf{Task}} & 
\makecell{\textbf{Baseline Reward}\\$\boldsymbol{\mu\!\pm\!\sigma}$} & 
\makecell{\textbf{GCR-PPO Reward}\\$\boldsymbol{\mu\!\pm\!\sigma}$} & 
\makecell{\textbf{Max Reward}\\(\textbf{Baseline}\,$\mid$\,\textbf{Ours})} & 
\textbf{SPC (\%)} & 
\makecell{\textbf{Avg.}\\\textbf{conflict}} \\
\midrule

Quadcopter &
\ms{64.205}{1.610} &
\ms{\textbf{64.911}}{\textbf{1.583}} &
67.126\, $\mid$\, 66.446 &
\textbf{1.09} &
0.15 \\

Reach-UR10 & 
\ms{0.740}{0.006} & 
\ms{\textbf{0.741}}{\textbf{0.016}} & 
0.747\,$\mid$\,0.765 & 
\textbf{0.13} & 
0.18 \\

Reach-Franka & 
\ms{\textbf{0.749}}{\textbf{0.011}} & 
\ms{0.743}{0.013} & 
0.759\,$\mid$\,0.766 & 
-0.80 & 
0.2 \\

Repose-Cube-Shadow & 
\ms{\textbf{9452.69}}{\textbf{205.21}} & 
\ms{9230.76}{410.37} & 
9759.18\,$\mid$\,9906.31 & 
-2.36 & 
0.39 \\

Velocity-Rough-Unitree-Go2 & 
\ms{\textbf{36.84}}{\textbf{0.25}} & 
\ms{36.34}{0.65} & 
37.23\,$\mid$\,37.35 & 
-1.37 & 
    0.45 \\

Lift-Cube-Franka & 
\ms{157.25}{4.54} & 
\ms{\textbf{158.83}}{\textbf{0.63}} & 
159.65\,$\mid$\,159.85 & 
\textbf{1.00} & 
0.91 \\

Velocity-Rough-H1 & 
\ms{35.54}{0.14} & 
\ms{\textbf{35.56}}{\textbf{0.34}} & 
35.71\,$\mid$\,36.01 & 
\textbf{0.06} & 
1.2 \\

Open-Drawer-Franka & 
\ms{98.73}{1.03} & 
\ms{\textbf{98.98}}{\textbf{0.63}} & 
99.84\,$\mid$\,99.70 & 
\textbf{0.25} & 
2.32 \\

Ant & 
\ms{195.12}{10.13} & 
\ms{\textbf{197.48}}{\textbf{14.24}} & 
208.67\,$\mid$\,214.48 & 
\textbf{1.20} & 
2.49 \\

Humanoid & 
\ms{276.58}{27.25} & 
\ms{\textbf{329.79}}{\textbf{24.79}} & 
309.08\,$\mid$\,355.73 & 
\textbf{19.24} & 
3.13 \\

Velocity-Rough-G1 & 
\ms{46.47}{1.18} & 
\ms{\textbf{48.10}}{\textbf{1.66}} & 
48.50\,$\mid$\,50.58 & 
\textbf{3.47} & 
3.23 \\

Repose-Cube-Allegro & 
\ms{167.70}{6.94} & 
\ms{\textbf{174.81}}{\textbf{5.91}} & 
178.99\,$\mid$\,183.02 & 
\textbf{4.12} & 
4.55 \\

Tracking-LocoManip-Digit & 
\ms{23.45}{3.78} & 
\ms{\textbf{33.06}}{\textbf{11.36}} & 
29.23\,$\mid$\,49.97 & 
\textbf{34.20} & 
30.71 \\

\bottomrule

\end{tabular}
}

\end{table*}

\subsection{IsaacLab Task Suite}\label{eval:isaac}
\begin{figure}[t!]
    \centering
    \includegraphics[width=1\linewidth]{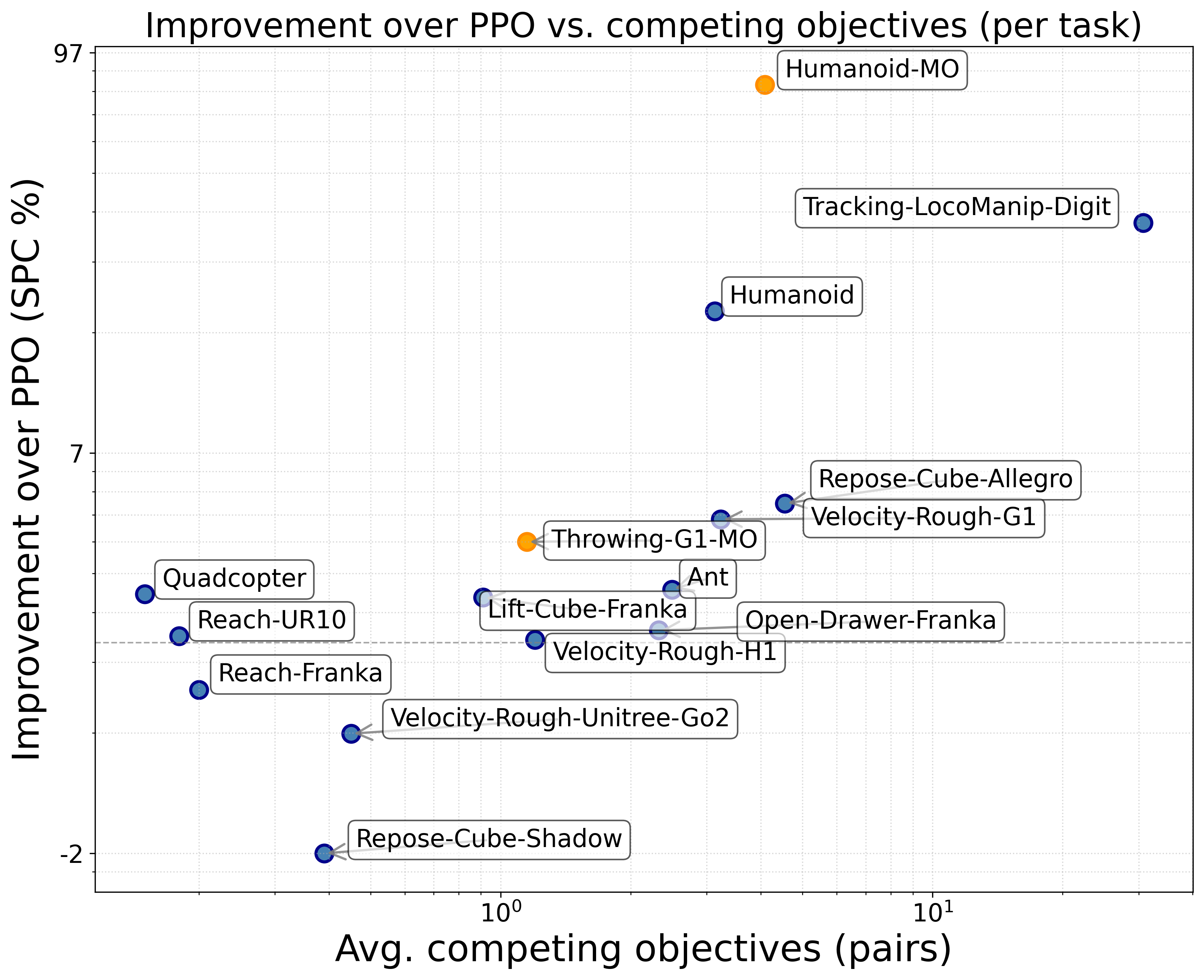}
    \caption{Relative improvement (symmetric percentage change) of GCR-PPO over PPO, averaged across 10 seeds each, as a function of the average number of competing objective pairs per task. Blue points are standard IsaacLab tasks; orange points are our custom multi-objective variants (Section~\ref{eval:multitask}).}
    \label{fig:scaling}
    \vspace{-0.5cm}
\end{figure}

Thirteen tasks were selected for evaluation, spanning both manipulation and locomotion across a range of robot platforms. We omitted some tasks because they were redundant with others (e.g., multiple quadruped walking tasks) or lacked a reward definition. Table \ref{tab:task-comparison} reports the returns and conflict levels for the selected tasks. Conflict was measured as the number of objective gradient conflicts ($>$90$^\circ$) through training, with the average shown. Task descriptions are available in \cite{IsaacLabDocs2025}.

During training, we observed differences in policy entropy between the baseline and our method when the entropy coefficient was not re-tuned, as our approach tended to decrease the learned standard deviation $\sigma$ of the actor's Gaussian action distribution. We tuned the entropy coefficient $\lambda$ separately for each method and task to address this. Specifically, we sampled $\lambda$ on a log scale from $[10^{-4}, 0.03]$, evaluated each over three seeds for half the training iterations, and selected the value yielding the highest reward. All other hyperparameters followed the defaults of the original task environments. In rare cases ($\leq$2\%), numerical instabilities during tuning and in the final training runs occurred in both the baseline and GCR-PPO, producing abnormally low rewards. These runs were rerun with new random seeds when the z-score exceeded three standard deviations.

The average symmetric percentage change (SPC) across tasks was 4.55\% (Table \ref{tab:ablation_spc}), increasing to 9.5\% when including the multi-objective tasks in Section \ref{eval:multitask}. SPC is defined as $SPC(a,b) = \frac{b-a}{\tfrac{1}{2}(a+b)}$, and is used here because its symmetric form avoids biasing reward increases over decreases compared with standard percentage change. However, SPC alone is limited, since most tasks are unbounded and lack explicit reward ranges. We therefore also consider win rates. Using paired t-tests with 95\% confidence, GCR-PPO achieves statistically significant improvements on Humanoid ($p = 2.4 \times 10^{-4}$), Velocity-Rough-G1 ($p = 0.022$), Repose-Cube-Allegro ($p = 0.024$), and Tracking-LocoManip-Digit ($p = 0.028$). Velocity-Rough-Unitree-Go2 shows a significant drop ($p = 0.043$). Variance across seeds is relatively high compared to mean returns. Overall, GCR-PPO outperforms PPO in 10 of 13 tasks, with a Bernoulli test giving $p = 0.046$ for this win rate.

Figure~\ref{fig:scaling} shows SPC improvements plotted against average gradient conflict per task. Across the IsaacLab tasks, the Spearman rank correlation is 0.736 ($p = 0.0041$), indicating a strong monotonic association. Conflict is measured as the average number of objective pairs with angles greater than 90° over all training steps and seeds. This metric directly quantifies the conflicts that GCR-PPO resolves and naturally increases with the number of reward terms. However, we use it rather than simply counting reward components, since tasks with the same number of terms can exhibit very different levels of conflict. We do not normalise by the number of objectives, since doing so would downplay the added difficulty of resolving conflicts as the number of objectives increases.

\renewcommand{\arraystretch}{1.1}
%

\subsection{Custom Multi-Objective Benchmarks}\label{eval:multitask}

\begin{figure}[!t]
    \centering
    \includegraphics[width=0.9\linewidth]{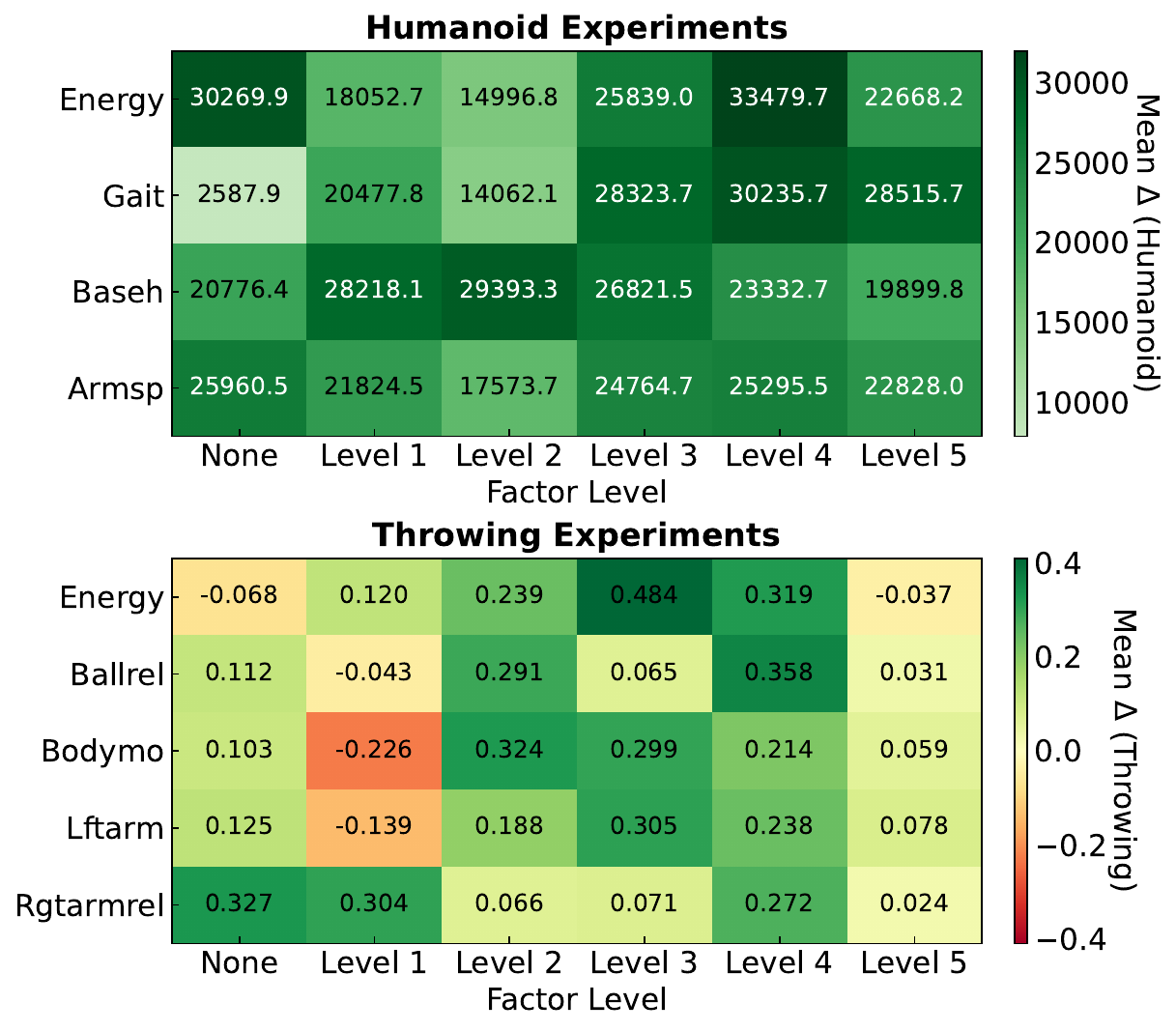}
    \caption{Performance difference (final reward) between GCR-PPO and PPO (massively GPU-distributed) on two custom multi-objective tasks: Humanoid Running and Full-Body Throwing. Results are averaged over reward ranges for each added objective to highlight the effect of task combinations on performance. Green indicates improvement over PPO, while red indicates a decrease. "None" denotes objectives not tested in a given experiment.}
    \label{fig:multiobjplt}
    \vspace{-0.5cm}
\end{figure}

\begin{table}[t]
\centering
\scriptsize
\setlength{\tabcolsep}{2pt} 
\caption{Win rates (\%) of GCR-PPO over PPO across factor levels for Full-Body Throwing. $n$ = configs.}
\resizebox{\columnwidth}{!}{%
\begin{tabular}{lcccccc}
\toprule
Factor & L0 & L1 & L2 & L3 & L4 & NA \\
\midrule
ENERGY  & 70.6 (17) & 80.0 (10) & 90.0 (10) & 100.0 (14) & 84.6 (13) & 54.3 (35) \\
BASEH   & 57.1 (14) & 90.0 (10) & 85.7 (7)  & 92.3 (13) & 90.9 (11) & 63.6 (44) \\
BALLREL & 72.7 (11) & 66.7 (12) & 88.2 (17) & 66.7 (9)  & 69.2 (13) & 73.0 (37) \\
BODYMO  & 84.6 (13) & 75.0 (12) & 81.2 (16) & 73.3 (15) & 62.5 (8)  & 68.6 (35) \\
LFTARM  & 76.9 (13) & 58.3 (12) & 72.7 (11) & 66.7 (12) & 80.0 (15) & 77.8 (36) \\
RGTARM  & 90.0 (10) & 72.7 (11) & 84.6 (13) & 58.3 (12) & 80.0 (10) & 69.8 (43) \\
\bottomrule
\end{tabular}%
}
\label{tab:throwing_winrates}
\end{table}

\begin{table}[t]
\centering
\scriptsize
\setlength{\tabcolsep}{2pt}
\caption{Win rates (\%) of GCR-PPO over PPO across factor levels for Humanoid Running. $n$ = configs.}
\resizebox{\columnwidth}{!}{%
\begin{tabular}{lcccccc}
\toprule
Factor & L0 & L1 & L2 & L3 & L4 & NA \\
\midrule
ENERGY & 100 (9)  & 100 (12) & 100 (11) & 100 (9)  & 100 (9)  & 100 (24) \\
GAIT   & 100 (6)  & 100 (13) & 100 (10) & 100 (6)  & 100 (16) & 100 (23) \\
BASEH  & 100 (12) & 100 (5)  & 100 (8)  & 100 (15) & 100 (8)  & 100 (26) \\
ARMSP  & 100 (10) & 100 (8)  & 100 (5)  & 100 (7)  & 100 (13) & 100 (31) \\
\bottomrule
\end{tabular}%
}
\label{tab:humanoid_winrates}
\vspace{-0.6cm}
\end{table}


Because the IsaacLab benchmarks are strongly tailored to PPO through extensive tuning and reward shaping, we designed additional tasks to provide a more challenging evaluation. We selected two settings: \textbf{Full-Body Throwing} \cite{munn2024whole} and \textbf{Humanoid Running} \cite{IsaacLabDocs2025}. We chose these because (i) they represent distinct whole-body control problems, and (ii) (ii) both occur on flat ground, which simplifies the design of style-based reward components. For each task, additional objectives were defined as threshold-based rewards: the agent received a fixed bonus whenever its behaviour satisfied the specified condition at a given time step. For example, a reward was granted whenever the robot’s hip remained within a target height range. This formulation ensured equal weighting across objectives, avoiding issues with continuous reward magnitudes. To increase diversity, the reward thresholds for each objective were partitioned into five non-overlapping ranges sorted in ascending order, with one range sampled per experiment. For each target number of objectives, 20 random subsets were drawn, up to a maximum of four for \textbf{Humanoid Running} and six for \textbf{Full-Body Throwing}. This provided a suitable set of tasks that would test the robustness of each method.

We extended the \textbf{Full-Body Throwing} and \textbf{Humanoid Running} tasks with additional objectives to increase their multi-objective complexity. For \textbf{Full-Body Throwing} (Throwing-MO), six were added: \textit{energy} (joint power consumption), \textit{baseh} (hip height), \textit{ballrel} (timing of ball release), \textit{bodymo} (torso velocity at release), and \textit{lftarm} and \textit{rgtarm} (hand heights at release). The implementation follows \cite{munn2024whole} using a Unitree G1 humanoid, with three additional penalties to regularise the policy toward realistic behaviour: action rate (scale $-1 \times 10^{-3}$), \texttt{dof\_torques\_l2} (scale $-2.5 \times 10^{-6}$), and \texttt{dof\_acc\_l2} (scale $-2.5 \times 10^{-8}$). For additional task objectives, scales are set to $1$ for objectives rewarded once per episode and $\texttt{step\_dt}$ for those rewarded at every step. For \textbf{Humanoid Running} (Humanoid-MO), four were introduced: \textit{energy} (scale $=5$), \textit{gait} (stride length scaled by contact duration, scale $=250$), \textit{baseh} (scale $=5$), and \textit{armsp} (arm–pelvis distance, scale $=5$). Each objective was defined through five non-overlapping ranges, with one range randomly selected per experiment. Agents received a fixed reward when their behaviour fell within the active range, ensuring equal weighting across objectives and avoiding bias from continuous shaping. Reward scales were set so that the reward magnitude was roughly proportional to the percentage of objective satisfaction, with additional objectives contributing at most a combined twofold increase over the original reward.

\begin{figure}[!h]
    \centering
    \includegraphics[width=1\linewidth]{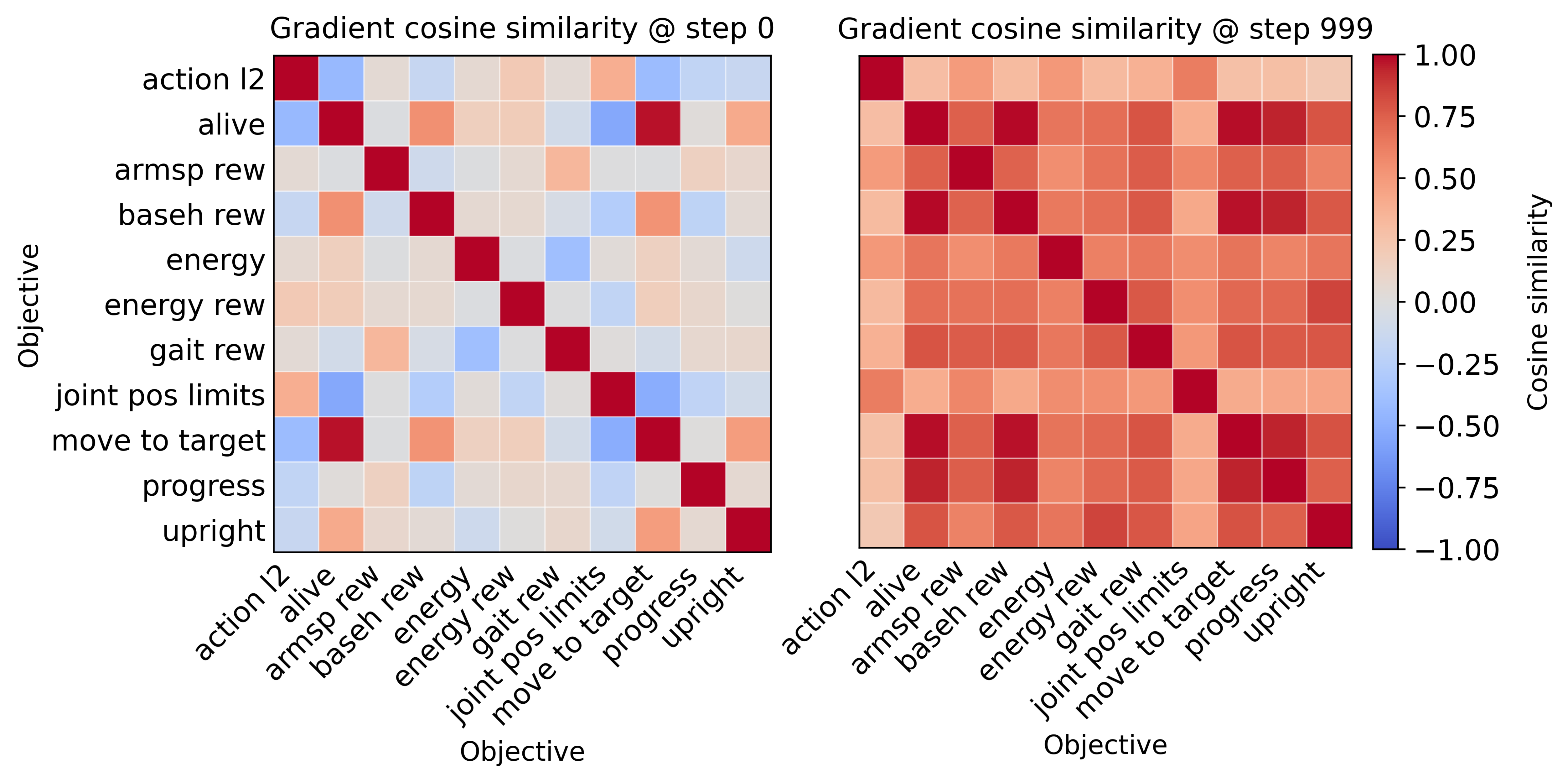}
    \caption{Conflict evolution in the Humanoid Running custom task, measured by cosine similarity between component gradients: left at training start, right at training end. With GCR-PPO, gradient conflicts decrease over training.}
    \label{fig:cosine}
\end{figure}

Figure \ref{fig:scaling} shows average conflict and SPC improvement for the custom \textbf{Humanoid Running} (79.7\% SPC, 4.09 average conflicts) and \textbf{Full-Body Throwing} (2.6\% SPC, 1.15 average conflicts) tasks. Both lie above the IsaacLab trend, with Humanoid Running a strong outlier (+74.8\% residual), likely due to the lack of reward shaping present in the IsaacLab benchmark. Within-task analyses found no significant correlation between conflict and SPC, as many objective ranges are mutually infeasible (e.g., minimising energy while maximising gait). When feasible, however, GCR-PPO reduces conflict over training time (Figure \ref{fig:cosine}).  

Figure~\ref{fig:multiobjplt} and Tables~\ref{tab:throwing_winrates}-\ref{tab:humanoid_winrates} show returns and win-rates, with all combinations exceeding 50\%. Performance depends on objective ranges: infeasible ones (e.g., minimal energy in Throwing) hurt performance, while feasible ones (e.g., moderate base height, sufficient momentum, large gait) improve it. Some infeasible ranges (e.g., minimal-energy Humanoid Running) still outperform PPO, as GCR-PPO leverages other objectives, whereas PPO overfits the conflicting reward.

\subsection{Ablations}
We evaluate three variants against PPO: (i) a multi-head critic baseline that computes per-term GAE and surrogate losses, (ii) GCR-PPO-NoPriority, which treats all objectives equally, and (iii) GCR-PPO with task regulariser prioritisation. Results are shown in Table \ref{tab:ablation_spc}. The multi-head critic yields a modest 2.1\% average improvement on IsaacLab (7/13 tasks) and a 12.04\% drop on the custom tasks. GCR-PPO-NoPriority improves over PPO on IsaacLab (4.8\% SPC vs. 4.6\% for GCR-PPO), but performs poorly on custom tasks—30.5\% improvement on Humanoid Running versus 79.7\% for GCR-PPO, and a 65.0\% drop on Throwing. Its lower variance reflects policies solving the primary task but failing to satisfy added objectives. Averaging across all 15 tasks, GCR-PPO achieves the strongest overall performance.     

\begin{table}[h!]
\centering
\footnotesize
\setlength{\tabcolsep}{4pt}
\caption{Ablation: average improvement (SPC) by method on IsaacLab, custom multi-objective (MO), and overall.}
\begin{tabular}{lccc}
\toprule
Method & IsaacLab & Custom (Humanoid,Throwing) & Overall  \\
\midrule
Multi-head PPO         &                 2.1\% &                -12.04\% &               0.21\% \\
GCR-PPO-NoPriority    &                 \textbf{4.8\%} &                -17.25\% &               1.86\% \\
GCR-PPO               &                 4.6\% &                \textbf{41.2\%} &               \textbf{9.5\%} \\
\bottomrule
\end{tabular}
\label{tab:ablation_spc}
\vspace{-0.3cm}
\end{table}

\vspace{-0.3cm}
\subsection{Training Performance}

\begin{figure}[!h]
    \centering
    \includegraphics[width=1\linewidth]{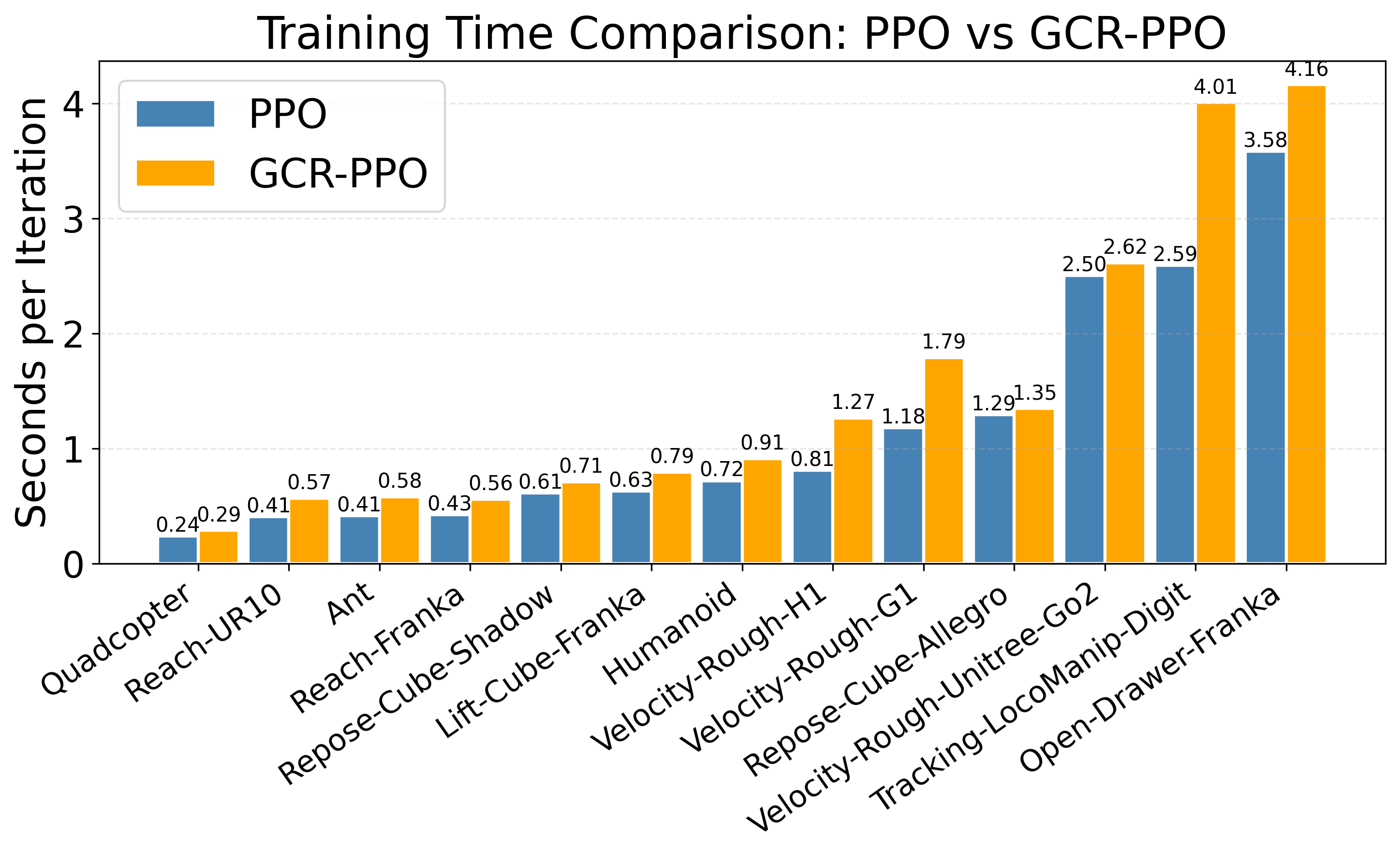}
    \caption{Time per training iteration (s) on a single NVIDIA H100 GPU, including data collection and model updates.}
    \label{fig:runtime}
    \vspace{-0.5cm}
\end{figure}

Figure~\ref{fig:runtime} shows iteration time for PPO and GCR-PPO on the IsaacLab benchmark. Nearly all overhead in GCR-PPO comes from gradient resolution (>99\%), while per-component loss calculation is negligible. Tasks with many reward components (e.g., Velocity-Rough-G1, Velocity-Rough-H1, Tracking-LocoManip-Digit), therefore, show a larger slowdown, with a worst case of 54\% ($\approx$ 47 minutes extra). The projection step scales linearly with conflicting pairs, though the number of potential pairs grows quadratically with reward terms. Overall, GCR-PPO delivers consistent improvements in return with only a modest increase in training time.

\section{CONCLUSIONS, DISCUSSION AND FUTURE WORK}

We presented GCR-PPO, an on-policy method that decomposes additive rewards into per-objective advantages and resolves conflicts via projection with task–regulariser prioritisation. Across 13 IsaacLab tasks and two custom multi-objective benchmarks, GCR-PPO improves returns and win rates over PPO, with larger gains in settings exhibiting higher conflict. The method integrates easily into PPO pipelines with modest training-time overhead.

Results highlight that some of the difficulty in multi-objective control stems from hidden interference between reward components. Exposing per-objective signals and applying lightweight conflict resolution improves stability without costly multi-objective solvers. Limitations include increased seed variability and added training time from the gradient surgery step. Our priority scheme encodes designer intent but does not provide Pareto guarantees.

Future work includes: (i) analysing why and where conflicts persist; (ii) extending to multiplicative or non-linear rewards; (iii) improving efficiency of gradient resolution and with more principled gradient resolution methods; and (iv) developing harder, less-shaped benchmarks. 

In summary, GCR-PPO shows that explicit reward geometry and resolving conflicts inside PPO are practical steps toward scalable multi-objective robot learning, with opportunities to generalise to richer objectives and real-world scenarios.







\bibliography{references}

\end{document}